\documentclass{article}

\usepackage{microtype}
\usepackage{graphicx}
\usepackage{subfigure}
\usepackage{booktabs} 

\usepackage[nohyperref,accepted]{icml2025}

\usepackage{amsmath}
\usepackage{amssymb}
\usepackage{mathtools}
\usepackage{amsthm}

\usepackage[utf8]{inputenc}
\usepackage[LAE,T1]{fontenc}
\usepackage{arabtex}
\usepackage{utf8}

\newcommand{\ku}[1]{%
    \mbox{\setcode{utf8}\RL{#1}}
}

\usepackage{times}
\usepackage{mathptmx}
\usepackage{newtxtext,newtxmath}

\usepackage[capitalize,noabbrev]{cleveref}

\theoremstyle{plain}

\theoremstyle{definition}

\theoremstyle{remark}

\begin{document}

\twocolumn[
\icmltitle{Subword Tokenization Strategies for Kurdish Word Embeddings}



\icmlsetsymbol{equal}{*}

\begin{icmlauthorlist}
\icmlauthor{Ali Salehi}{yyy}
\icmlauthor{Cassandra L. Jacobs}{yyy}
\end{icmlauthorlist}

\icmlaffiliation{yyy}{Department of Linguistics, University at Buffalo, Buffalo NY, USA}

\icmlcorrespondingauthor{Ali Salehi}{asalehi@buffalo.edu}

\icmlkeywords{Machine Learning, ICML}

\vskip 0.3in
]
\printAffiliationsAndNotice{ }  

\begin{quote}
We investigate tokenization strategies for Kurdish word embeddings by comparing word-level, morpheme-based, and BPE approaches on morphological similarity preservation tasks. We develop a BiLSTM-CRF morphological segmenter using bootstrapped training from minimal manual annotation and evaluate Word2Vec embeddings across comprehensive metrics including similarity preservation, clustering quality, and semantic organization.
Our analysis reveals critical evaluation biases in tokenization comparison. While BPE initially appears superior in morphological similarity, it evaluates only 28.6\% of test cases compared to 68.7\% for morpheme model, creating artificial performance inflation. When assessed comprehensively, morpheme-based tokenization demonstrates superior embedding space organization, better semantic neighborhood structure, and more balanced coverage across morphological complexity levels. These findings highlight the importance of coverage-aware evaluation in low-resource language processing and offers different tokenization methods for low-resourced language processing.
\end{quote}



\vspace{-.6cm}

\section{Introduction}
Effective word representations are critical for natural language processing, particularly for low-resource, morphologically rich languages where data scarcity compounds the complexity of linguistic structures \citep{Erdmann2018ComplementarySFA, rudersurevey}. 
Tokenization, the process of segmenting text into units, is the foundation for these representations, with significant impact on downstream applications. 
Conventional word-level tokenization approaches fail to capture the compositional nature of meaning encoded in morphological structures in languages with rich morphology \cite{cotterell2016morphological}. 
While linguistic theory suggests morpheme-based tokenization should outperform statistical approaches for morphologically rich languages \cite{luong2013better, Park2020MorphologyMAA}, some empirical evidence has begun to challenge this assumption. 
Recent research across various languages has revealed a surprising gap between theoretical expectations and practical results in subword tokenization \cite{bostrom2020byte}. 
Statistical methods like Byte-Pair Encoding (BPE) \cite{sennrich2016neural}, which merge frequent character sequences without linguistic guidance, sometimes outperform linguistically-informed approaches \cite{mielke2019kind}. 
On the other hand, comparative research on morphologically rich languages like Turkish, Finnish, and Hungarian \cite{cotterell2016morphological, creutz2005unsupervised} has demonstrated the benefits of morphologically-informed representations.

In this work, we present an in-depth analysis of the importance of tokenization for Kurdish natural language processing due to its complex morphology \cite{esmaili2013sorani}. 
Despite its historical and cultural significance, Kurdish remains under-resourced in computational linguistics \cite{hassani2018blark}, with limited exploration of optimal tokenization strategies. 
The unexpected findings that subword tokenization is sufficient for most tasks necessitates systematic evaluation of different segmentation strategies.
Here we ask which tokenization approaches are most optimal for Kurdish to narrow the gap in the literature \cite{ahmadi2019tokenization}. 
We review the linguistic properties of Kurdish and present a comparison of word-level, morpheme-based, and subword tokenization strategies for Kurdish, examining their impact on word embedding quality and downstream tasks. 
Our work addresses the crucial need for empirically-grounded tokenization approaches in low-resource settings \cite{gerz2018relation} and contributes to the broader understanding of representation learning for morphologically complex languages. 
    
\section{Challenges for Kurdish Language Processing}

Kurdish morphology exhibits extensive derivational and inflectional processes \cite{thackston2006sorani}. Nouns encode definiteness and number, while verbs express tense, aspect, mood, person, and number through complex affixation patterns. 
The agglutinative structure of Kurdish particularly complicates verbal constructions where multiple morphemes concatenate to form complex predicates \cite{ahmadi2020towards}. For example, the Kurdish verb \textit{neyandegerandewe} (they wouldn't return it) contains six morphemes: \textit{ne-} (negation) + \textit{yan} (3rd person plural object) + \textit{de-} (past imperfective) + \textit{gerênd} (causative `return') + \textit{ewe} (directional `back').
The morphological density of Kurdish words is comparable to other morphologically rich languages and creates compositional structures where single words encode multiple features. 
The noun \ku{کتێبەکانم} (\textit{kitêbekanim}, ``my books") demonstrates typical complexity: \ku{کتێب} (\textit{kitêb}, ``book") + \ku{ەکان} (\textit{ekan}, plural) + \ku{م} (\textit{im}, possessive).
This morphological complexity yields high type-to-token ratios, exacerbating data sparsity in computational models.

In addition to its morphological properties, Kurdish lacks the large-scale annotated corpora and computational lexicons available for high-resource languages \cite{hassani2018blark}, though large, unlabeled datasets are increasingly available.
Dialectal variation and sociolinguistic fragmentation adds complexity, with Sorani and Kurmanji differing in morphology, syntax, and vocabulary \cite{sheykhsorani}. 
Orthographic inconsistencies arise from optional vowel representation in the Arabic-derived script, where short vowels may be omitted \cite{ahmadi-2020-tokenization}.
Morphological analysis tools have developed incrementally, with notable efforts including Sorani analyzers \cite{Ahmadi2020TowardsFM, Ahmadi2021HunspellFS}, finite-state transducers for Kurdish \cite{Ahmadi2020TowardsFM}, and the comprehensive AsoSoft framework \cite{asosoftcorpuspaper, MAHMUDI2021101222}, which provides large text collections, transliteration systems, and web-accessible morphological analysis. These tools achieve reasonable coverage for standard texts but struggle with neologisms, borrowed terminology, and social media content with spelling variations.

We argue here that linguistically informed tokenization that combines the strengths of statistical methods like subword segmentation \cite{sennrich2016neural} and morphological boundary detection presents a potential solution for the orthographic and morphophonemic alternations of Kurdish.
To address these issues, we present a neural BiLSTM-CRF model that addresses these limitations through bootstrapped training from minimal annotation (1,540 words), demonstrating effective morphological analysis with limited resources.

\section{Methodology}

The training process for our morphological segmentation model followed a bootstrapping approach. 
Our study utilized the AsoSoft Text Corpus  \citep{asosoftcorpuspaper} as the primary data source, which is one of the largest available collections of Kurdish (Sorani) text. 
We manually segmented approximately 1,500 Kurdish words based on linguistic morphological analysis. These words were randomly selected from the corpus to ensure coverage of different word types and morphological patterns including different parts of speech,  light verb constructions, preverbal constructions and compounds. These words served as the initial training set for the BiLSTM-CRF model. After training this initial model, we applied it to segment additional words from our corpus, manually verified a subset of these new segmentations, and added them to our training data. Through this iterative process, we expanded our training set to over 4,000 words with gold-standard morphological segmentations. 

\subsection{Text normalization}

Text normalization for Kurdish presented numerous challenges due to its non-standardized orthography, dialectal diversity, and the nature of the available corpus. 
Given the lack of standard Kurdish NLP preprocessing libraries, every aspect of the workflow had to be developed from scratch or extensively adapted. 
Our preprocessing pipeline was developed across several stages
, requiring considerable manual tuning and verification. 
We applied AsoSoft's text normalization method \citep{MAHMUDI2021101222, asosoftcorpuspaper} from their Python library in the initial preprocessing stages to address standard Kurdish text inconsistencies before implementing our custom normalization procedures.
This multi-phase normalization required several iterations of testing, reviewing, and modifying rules across millions of tokens. 
The final preprocessed corpus formed the foundation of our tokenization and embedding experiments, and its quality was critical to the validity of all subsequent results.

The Asosoft corpus is constructed from various sources, with news articles constituting the majority of the content. 
After extensive preprocessing, our final cleaned corpus contained 24.5 million tokens spanning approximately 2.3 million sentences.
We first defined a strict set of allowable characters that included letters from the extended Arabic-based script used in Sorani Kurdish, numerals, and a limited range of punctuation marks. All non-Kurdish or extraneous characters were filtered using regular expressions. This filtering process was not straightforward, as informal writing styles, character borrowing from Persian and Arabic, and inconsistent Unicode encodings introduced significant noise in the text (see Appendix).

Corpus cleanup also included extensive deduplication, including exact sentence duplicates and fuzzy duplicates using token-overlap measures. We removed near-identical headlines, repeated paragraphs, and templated sentences across documents. Sentences that fell below a minimum token threshold or lacked valid word structure were discarded (see appendix).
In the final phase of preprocessing, we implemented Kurdish-specific sentence segmentation rules to extract clean sentence boundaries using heuristics around punctuation and spacing. Given the inconsistent use of sentence delimiters and overlap with non-Kurdish scripts, this step required custom filtering to remove embedded Persian and Arabic segments. The resulting corpus of 2.3 million well-formed sentences was further deduplicated and formatted into sentence-per-line and word-per-line variants to support downstream tokenization schemes.

\subsection{Tokenization approaches}

Tokenization strategies for morphologically rich languages fundamentally shape how models represent linguistic structure, spanning from statistical to linguistically-informed methods. This comparison addresses a core theoretical question: whether linguistically-informed segmentation outperforms statistical frequency-based approaches for capturing meaningful morphological relationships in Kurdish. We quantify the convergence between these approaches through segmentation agreement analysis, measuring boundary alignment using similarity coefficients.


Statistical approaches like Byte-Pair Encoding (BPE; \citealp{sennrich2016neural}) operate through iterative merging, starting with characters and incrementally merging frequent adjacent pairs $p = (x, y)$ using $\arg\max_{p \in V} \mathrm{count}(p)$ until reaching target vocabulary size. Alternative methods include WordPiece \cite{schuster2012japanese}, which incorporates likelihood criteria, and SentencePiece \cite{kudo2018sentencepiece}, which treats whitespace as regular characters. The unigram language model \cite{kudo2018subword} employs a different paradigm, starting with large vocabulary and iteratively removing subwords to maximize corpus likelihood. Unsupervised morphological approaches include Morfessor \cite{smit-etal-2014-morfessor}, which applies minimum description length principles to automatically discover morpheme boundaries. This method seeks segmentations that minimize combined encoding costs of both lexicon (morpheme inventory) and corpus (word occurrences), balancing between poor corpus compression (too few morphemes) and excessive lexicon size (too many morphemes).

By contrast, linguistically-motivated segmentation identifies meaningful units $s$ aligned with linguistic structures such as words ($w$) rather than frequency patterns, formulated as:
\begin{equation}
\arg\max_{s_1, s_2, ..., s_n} P(s_1, s_2, ..., s_n | w)
\end{equation}
where the probability function incorporates morphological knowledge about valid combinations and morphotactic constraints.

\subsubsection{Morphological segmentation with BiLSTM-CRF}

For morpheme-level tokenization, we developed a BiLSTM-CRF neural architecture \citep{lafferty2001conditional, Huang2015BidirectionalLM} that predicts morpheme boundaries within Kurdish words. This approach decomposes complex forms into constituent morphemes, enabling capture of morphological regularities and improved generalization across paradigms. Given Kurdish's limited annotated data, we investigate whether effective segmentation is achievable through bootstrapping from minimal training data.

\subsubsection{Model Architecture}
The BiLSTM-CRF model processes words at the character level using an embedding-based approach. Our implementation begins with a character embedding layer that maps each character to a dense vector representation:
\begin{equation}
\mathbf{x}_t = \text{Embedding}(c_t) \in \mathbb{R}^{d}
\end{equation}
where $d$ is the embedding dimension for each character in the Kurdish alphabet.
The architecture consists of three main components: a multi-layer bidirectional LSTM that processes character sequences in both directions to capture contextual information, a linear projection layer that maps LSTM outputs to boundary prediction scores, and a Conditional Random Field (CRF) layer that enforces valid boundary label sequences.The model computation proceeds as follows:

\begin{align}
\mathbf{h}^{f}, \mathbf{h}^{b} &= \text{BiLSTM}(\mathbf{x}_{1:n}) \\
\mathbf{h}_t &= [\mathbf{h}_t^{f}; \mathbf{h}_t^{b}] \\
\mathbf{e}_t &= \mathbf{W}\mathbf{h}_t + \mathbf{b}
\end{align}

The CRF layer models the conditional probability of label sequence $\mathbf{y}$ given input $\mathbf{x}$:
\begin{equation}
P(\mathbf{y}|\mathbf{x}) = \frac{\exp(s(\mathbf{x},\mathbf{y}))}{\sum_{\mathbf{y}' \in \mathcal{Y}(\mathbf{x})}\exp(s(\mathbf{x},\mathbf{y}'))}
\end{equation}
where the score function combines emission and transition scores:
\begin{equation}
s(\mathbf{x},\mathbf{y}) = \sum_{t=1}^{n}[\mathbf{e}_{t,y_t} + \mathbf{T}_{y_{t-1},y_t}]
\end{equation}
Here, $\mathbf{h}^{f}$ and $\mathbf{h}^{b}$ represent the forward and backward LSTM hidden states, $\mathbf{e}_t$ are emission scores, and $\mathbf{T}$ is the transition score matrix.

\subsubsection{Training Configuration and Bootstrapping}
We employed a bootstrapping approach to train the model with minimal annotation resources. Beginning with 1,540 manually segmented Kurdish words, we iteratively expanded our training set through model-assisted annotation and manual verification, ultimately reaching over 4,000 annotated words. The model used a hidden size of 256 dimensions, 3 BiLSTM layers, and a dropout rate of 0.3 for regularization. We used the Adam optimizer with learning rate 0.001 and weight decay 1e-5, employing early stopping with 10 epochs patience to prevent overfitting.

We experimented with two labeling schemes: the ``end-only scheme'' where only the last character of each morpheme is marked as a boundary (0-0-1), and the ``both-ends scheme'' where both the first and last characters of each morpheme are marked as boundaries (1-0-1). After evaluation on our validation set, we selected the end-only scheme for our final model based on its superior performance.
During inference, the model segments words using the CRF's Viterbi decoding algorithm \citep{lafferty2001conditional} to find the optimal label sequence:

\begin{equation}
\mathbf{y}^* = \underset{\mathbf{y} \in \mathcal{Y}(\mathbf{x})}{\arg\max} \sum_{t=1}^{n}[\mathbf{e}_{t,y_t} + \mathbf{T}_{y_{t-1},y_t}]
\end{equation}

\subsubsection{Model Performance and Error Analysis}
The final model achieved an F1-score of .815 for boundary detection, with precision of .835 and recall of .796. However, performance evaluation revealed significant variation across word categories. Nouns achieved over 90\% segmentation accuracy and adjectives around 93\%, while verbs reached only 42\% accuracy. This discrepancy stems from Kurdish verb morphology's greater complexity, including extensive affixation patterns, irregular forms, single-character stems with phonological changes, ambiguous direction-marking affixes, and fewer training examples of complex verbal constructions.

After applying morphological segmentation to our full corpus, each word decomposed into an average of 1.99 morphemes, consistent with expectations for morphologically complex languages. This segmentation strategy significantly impacts embedding quality for morphologically related forms, particularly for nouns and adjectives where segmentation accuracy is high. We examine how this segmentation quality directly affects morphological similarity preservation in our embedding evaluation results (Section \ref{embedding-analyses}).

\subsubsection{Byte-Pair Encoding (BPE)}

We implemented Byte-Pair Encoding \citep{sennrich2016neural} as a tokenization approach that operates without explicit linguistic knowledge. BPE 
offers a data-driven alternative to morphologically-informed segmentation.

Our BPE implementation used the HuggingFace Tokenizers library with a target vocabulary size of 2,280 tokens based on empirical morpheme count analysis. This vocabulary size was chosen to create a compact representation that balances coverage and efficiency—small enough to ensure frequent occurrence of each subword unit for robust statistical learning, yet large enough to capture common Kurdish character sequences and morphological patterns. The training process included a minimum frequency threshold of 2 to filter out rare character combinations.

The resulting BPE tokenizer segments Kurdish words into subword units averaging 3.75 tokens per word, more granular than the morpheme-based approach (1.99 tokens per word) but more compact than character-level tokenization. For example, a complex Kurdish word like \ku{دەستپێکردنەوە} (``to start'') might be segmented as \ku{دەست-پێ-کرد-نەوە}, where the algorithm has learned to identify frequently co-occurring character sequences regardless of their linguistic significance.

\subsubsection{Word-level tokenization}

Word-level tokenization serves as our baseline approach, treating each word as a unit without internal decomposition. This traditional method employs whitespace and punctuation-based segmentation, creating the largest vocabulary among our three approaches. For Kurdish text processing, we applied standard tokenization rules while addressing language-specific considerations such as compound word boundaries and clitic attachment patterns. 

While word-level tokenization offers the advantage of preserving complete lexical meanings, it presents significant challenges for morphologically rich languages like Kurdish. The approach suffers from high data sparsity, as each inflected or derived form is treated as a distinct vocabulary item, and cannot generalize to unseen word forms. This limitation is particularly pronounced in Kurdish, where productive morphological processes can generate numerous variants of a base form, leading to substantial out-of-vocabulary issues and requiring extensive vocabulary coverage for adequate representation.

\subsection{Kurdish word, subword, and morpheme embeddings}

We employed the skip-gram word2vec architecture \cite{mikolov2013distributed} to train comparable embeddings across all three tokenization approaches. 
Our framework addresses the challenge of fair comparison between tokenization strategies that produce vastly different token counts per word. Our evaluation includes analysis of embedding space organization through separation ratios (measuring intra-lemma vs. inter-lemma clustering) and similarity dropoff rates (quantifying how similarity decreases across nearest neighbor rankings).


\subsubsection{Window Size Adjustment Methodology}

A critical part in our approach is the dynamic adjustment of context window sizes based on tokenization granularity. Since different tokenization strategies produce varying numbers of tokens per word, maintaining identical window sizes would create unfair comparisons where subword approaches see artificially truncated contexts.
We calculate the average tokens per word for each tokenization approach and adjust window sizes proportionally:
\begin{equation}
w_{\textit{adjusted}} = \lceil w_{\textit{base}} \times \text{avg\_tokens}_{\textit{approach}} \rceil
\end{equation}
where $w_{\textit{base}}$ represents the baseline window size for word-level tokenization.

\subsubsection{Training Configuration}

All models used identical training parameters: vector dimensions of 100 and 150, minimum count threshold of 5, 10 training epochs, and skip-gram with negative sampling (5 negative samples). 
Models were trained using Gensim's word2vec \citep{rehurek_lrec} implementation with consistent preprocessing pipelines to eliminate confounding factors. 
For out-of-vocabulary evaluation, morpheme and BPE models compute compositional vectors by averaging constituent subword embeddings, while the word-level model cannot handle unseen words.
Training consistency was ensured through fixed random seeds and multiple independent runs, with final results representing averaged performance across three training iterations.

\subsubsection{Handling Out-of-Vocabulary Words}

A critical consideration for fair evaluation is handling words not present in model vocabularies during testing. The three tokenization approaches exhibit fundamentally different capabilities for addressing out-of-vocabulary (OOV) words, which directly impacts evaluation coverage and comparability. Coverage refers to the percentage of evaluation word pairs for which a model can generate embeddings for both components.
We utilize coverage as the percentage of evaluation word pairs for which a model can generate embeddings for both the lemma and wordform, either through direct vocabulary lookup or compositional vector construction. Compositional vectors are created by averaging embeddings of constituent subword units when complete words are absent from the vocabulary.
Compositional vectors are constructed by averaging the embeddings of constituent subword units when the complete word is not present in the model vocabulary.

For morpheme and BPE models, we implement compositional vector generation by averaging constituent subword embeddings when the complete word form is absent from the vocabulary. When evaluating a word not present in the morpheme model, we decompose it using our BiLSTM-CRF segmenter and compute the mean vector of available morpheme embeddings. Similarly, BPE models leverage their learned subword units to construct representations for unseen words through vector averaging of constituent BPE tokens.
The word-level model, by design, cannot generate representations for OOV words, as it treats each word as an atomic unit. This limitation creates an inherent evaluation disadvantage, as word-level models have zero coverage for words absent from their training vocabulary. 

\section{Evaluation and Results}

Our evaluation methodology implements a framework designed to establish the relationship between segmentation quality and embedding performance while identifying distinct organizational patterns that different tokenization strategies create in embedding space. We present evaluation methodology alongside corresponding results to provide immediate insights into each analytical approach.

\subsection{Morphological Segmentation Quality Assessment}

We employ the UniMorph Kurdish dataset \cite{pimentel-ryskina-etal-2021-sigmorphon} as our gold standard for morphological segmentation evaluation. UniMorph provides morphologically annotated word forms with lemmas and feature annotations for about 1,000 Kurdish (Sorani) words, making it suitable for validating our BiLSTM-CRF segmentation accuracy.
We evaluate the BiLSTM-CRF segmentation system using F1-score calculation for binary boundary detection, treating each character position as either boundary (1) or non-boundary (0). Performance assessment includes systematic breakdown by part-of-speech categories, revealing substantial variation in segmentation difficulty across linguistic categories.

\begin{figure}[h]
\centering
\includegraphics[width=0.45\textwidth]{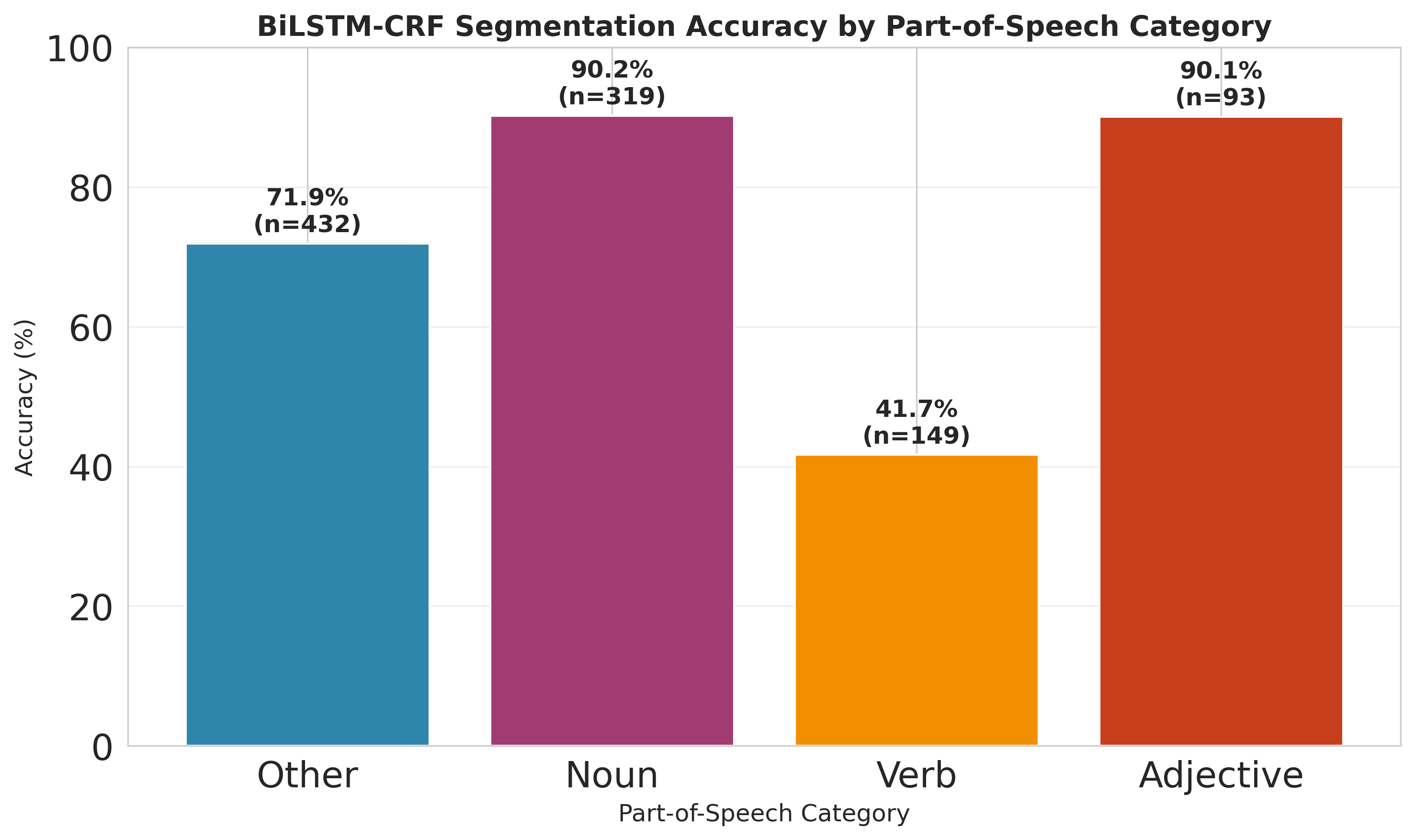}
\caption{BiLSTM-CRF morphological segmentation accuracy by part-of-speech category, showing substantial variation in boundary detection performance across linguistic categories.}
\label{fig:bilstm_pos_accuracy}
\end{figure}

Figure \ref{fig:bilstm_pos_accuracy} shows the substantial performance variation across POS categories detailed in our methodology.

\subsection{Morphological Similarity Analysis}

We evaluate how well each tokenization approach preserves morphological relationships by measuring cosine similarity between lemmas and their inflected forms from the UniMorph dataset:

\begin{equation}
\text{sim}(w_{\textit{lemma}}, w_{\textit{wordform}}) = \frac{v_{w_{\textit{lemma}}} \cdot v_{w_{\textit{wordform}}}}{||v_{w_{\textit{lemma}}}|| \cdot ||v_{w_{\textit{wordform}}}||}
\end{equation}

When complete words are absent from model vocabularies, morpheme and BPE models generate compositional vectors by averaging constituent subword embeddings, while word-level models cannot handle out-of-vocabulary cases.
BPE achieves the highest average similarity (0.73), followed by morpheme-based approaches (0.62) and word-level models (0.57). This unexpected result challenges the assumption that linguistically-informed segmentation inherently outperforms statistical methods.

\begin{table}[h]
\centering
\begin{tabular}{lc}
\toprule
\textbf{Approach} & \textbf{Similarity (±SD)} \\
\midrule
Word & 0.528 ± 0.15 \\
Morpheme & 0.583 ± 0.20 \\
BPE & 0.752 ± 0.18 \\
\bottomrule
\end{tabular}
\caption{Average morphological similarity scores with standard deviations}
\end{table}



BPE's apparently superior morphological similarity performance (0.752) must be interpreted cautiously due to severe evaluation coverage limitations. While BPE achieves higher average similarity with lower standard deviation (0.18), indicating more consistent scores, this performance is based on only 28.6\% of test cases compared to 94.3\% for word models.

The more structured embedding space organization demonstrated by morpheme models in neighbor rank analysis, combined with their superior clustering quality metrics, suggests that when fairly evaluated on comparable test sets, linguistically-informed segmentation may indeed outperform statistical approaches. The frequency-based patterns captured by BPE appear to create tighter morphological clustering for a limited subset of evaluable cases, but this advantage may not generalize to comprehensive morphological processing tasks.
This coverage bias represents a critical methodological limitation that necessitates restricted evaluation on mutually evaluable word sets to enable fair comparison between tokenization approaches.

\subsection{Segmentation Strategy Comparison}

We quantify alignment between morpheme and BPE segmentation boundaries using the Jaccard similarity coefficient \citep{manning1999foundations}. For each word $w$, we define $A$ as morpheme boundary positions and $B$ as BPE boundary positions:

\begin{equation}
\text{agreement}(w) = \frac{|A \cap B|}{|A \cup B|}
\end{equation}

This metric ranges from 0 (no shared boundaries) to 1 (perfect boundary alignment), enabling analysis of convergence between statistical and linguistic segmentation approaches.
Analysis reveals fundamental divergence between the two approaches. Despite processing identical text, morpheme and BPE methods achieve only 14.4\% average boundary agreement, with 63.6\% of words showing zero agreement and merely 2.5\% achieving perfect alignment. 

\begin{table}[h]
\centering
\small
\begin{tabular}{lcc}
\toprule
\textbf{Metric} & \textbf{Value} & \textbf{Count} \\
\midrule
Zero agreement & 63.6\% & 636 words \\
Perfect agreement & 2.5\% & 25 words \\
Average agreement & \multicolumn{2}{c}{14.4\%} \\
\midrule
Morpheme tokens/word & \multicolumn{2}{c}{1.99} \\
BPE tokens/word & \multicolumn{2}{c}{3.75} \\
\bottomrule
\end{tabular}
\caption{Segmentation agreement and tokenization density}
\label{tab:segmentation_agreement}
\end{table}

The predominance of zero-agreement cases (63.6\%) demonstrates that statistical and linguistic approaches identify almost entirely different sets of meaningful units. This fundamental divergence, combined with BPE's nearly doubled tokenization density, indicates that frequency-based patterns and morphological boundaries represent complementary rather than competing approaches to identifying linguistic structure. The minimal convergence suggests that hybrid methods combining both perspectives may be necessary to capture the full spectrum of meaningful units in Kurdish morphology.

\subsection{Embedding Space Organization Analysis}
\label{embedding-analyses}

We analyze embedding space structure through two complementary approaches: similarity distribution patterns and neighbor rank similarity analysis.

\subsubsection{Neighbor Rank Similarity Analysis}

\begin{figure}[h]
\centering
\includegraphics[width=0.45\textwidth]{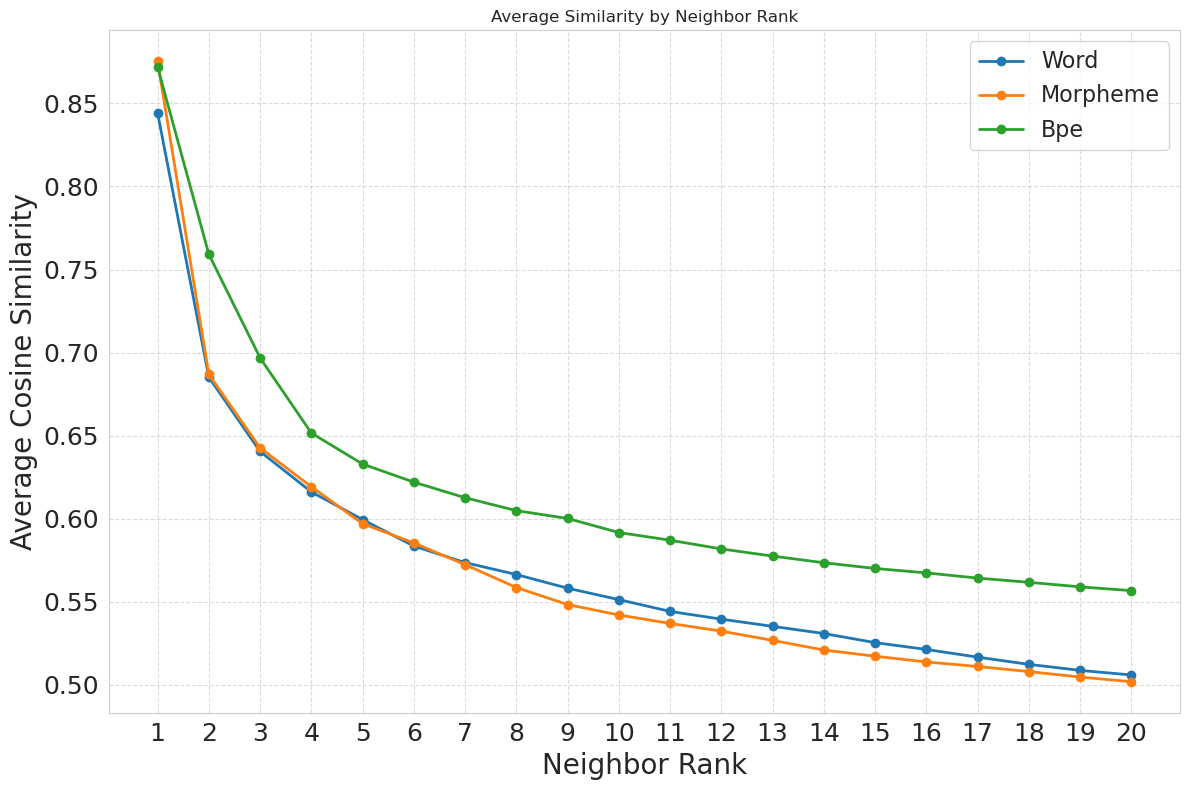}
\caption{Average similarity by neighbor rank, showing how similarity decreases across ranked nearest neighbors for each tokenization approach. BPE maintains higher similarities across all ranks while morpheme and word models show steeper decay patterns.}
\label{fig:similarity_dropoff}
\end{figure}

We examine how similarity decreases across ranked nearest neighbors by calculating average similarity at each neighbor rank (Figure \ref{fig:similarity_dropoff}). Steep dropoff indicates well-organized semantic clusters, while flat curves suggest poorly structured embedding spaces.
 Contrary to expectations, morpheme and word models show nearly identical dropoff patterns, while BPE demonstrates the \textit{shallowest dropoff} (i.e., flattest curve). All models start at similar similarity levels (~0.88-0.89) for rank 1 neighbors, but BPE maintains consistently higher similarities across all neighbor ranks, ending at ~0.54 at rank 20 compared to ~0.50 for morpheme and word models.

\begin{table}[h]
\centering
\begin{tabular}{lccc}
\toprule
\textbf{Dropoff Measure} & \textbf{Word} & \textbf{Morpheme} & \textbf{BPE} \\
\midrule
Rank 1-5 Dropoff & 31.2\% & 32.7\% & 30.3\% \\
Rank 1-10 Dropoff & 37.6\% & 38.8\% & 35.3\% \\
Rank 1-20 Dropoff & 43.1\% & 44.2\% & 39.1\% \\
\bottomrule
\end{tabular}
\caption{Similarity dropoff rates at neighbor rank intervals}
\label{tab:dropoff_rates}
\end{table}

The neighbor rank analysis reveals a counterintuitive pattern where BPE's  morphological similarity scores coincide with less structured embedding space organization. BPE's shallow dropoff (39.1\% rank 1-20) suggests more uniform semantic neighborhoods rather than the distinct clustering typically desired in embedding spaces. In contrast, morpheme and word models show steeper, more similar decay patterns (44.2\% and 43.1\% respectively), indicating better-defined semantic boundaries.

This finding challenges the interpretation of BPE's morphological similarity advantage, suggesting that higher similarity scores may reflect overly uniform embedding spaces rather than superior morphological understanding. The trade-off appears to favor consistency over discrimination: BPE creates embeddings where everything looks moderately similar to everything else, while morpheme-based approaches create more structured spaces with clearer semantic distinctions.

\subsubsection{Allomorph Clustering Analysis}

We analyze how different inflected forms of the same lemma cluster in embedding space by comparing intra-lemma distances (between forms of the same lemma) versus inter-lemma distances (between forms of different lemmas). This analysis reveals how well each tokenization approach groups morphologically related forms while maintaining separation between unrelated words.

\begin{figure}[h]
\centering
\includegraphics[width=.45\textwidth]{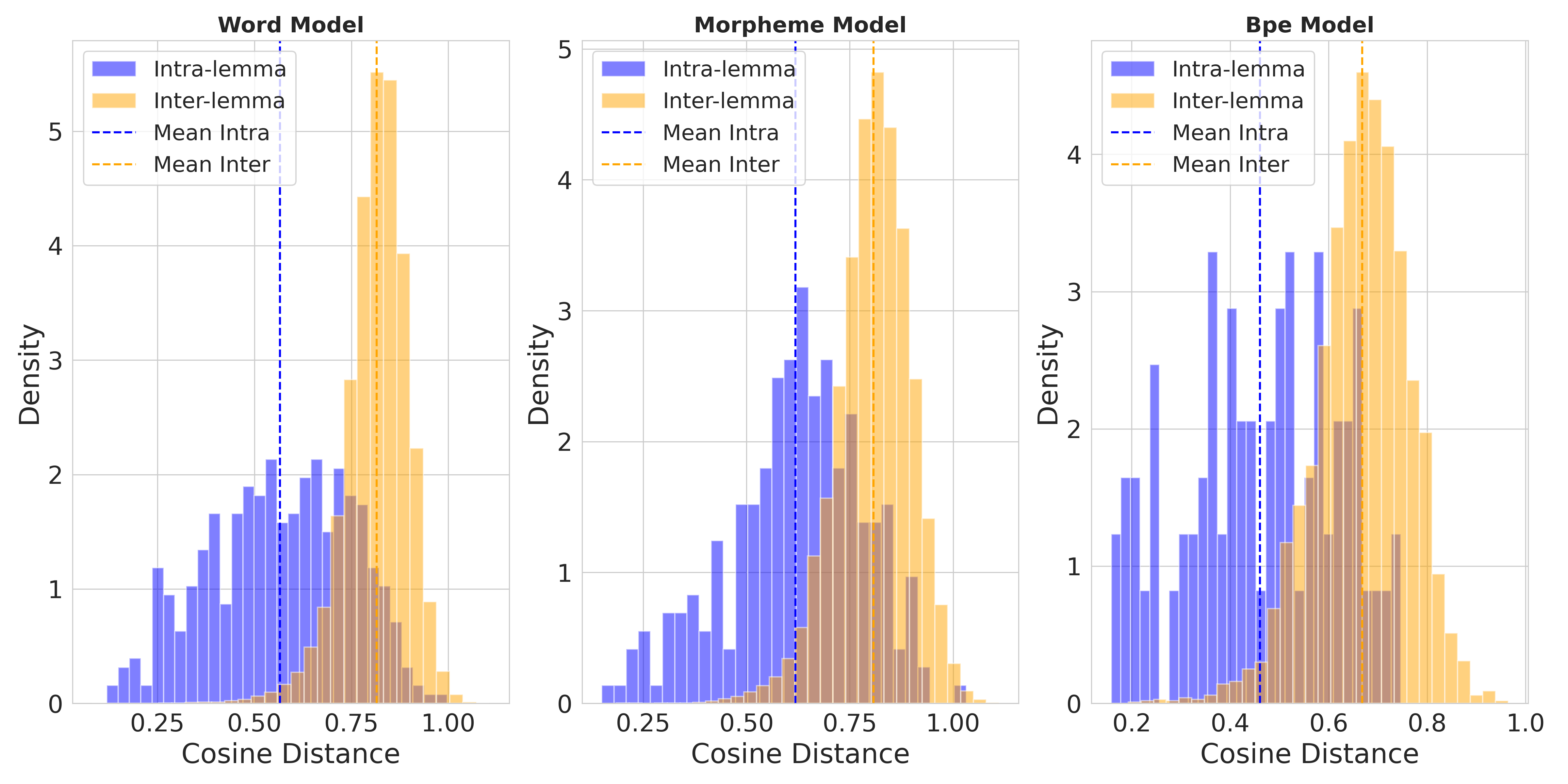}
\caption{Distribution of cosine distances for intra-lemma (same lemma) versus inter-lemma (different lemmas) word pairs across tokenization approaches. Blue histograms show distances between inflected forms of the same lemma, while orange histograms show distances between forms of different lemmas.}
\label{fig:allomorph_clustering}
\end{figure}

The distance distribution plots (Figure \ref{fig:allomorph_clustering}) reveal distinct clustering patterns across tokenization approaches. All models achieve similar separation ratios (Word: 1.44, Morpheme: 1.31, BPE: 1.45), calculated as the ratio of average inter-lemma distances to average intra-lemma distances, indicating comparable ability to distinguish between different lemmas while clustering related forms. However, the distribution shapes differ substantially. BPE demonstrates the tightest intra-lemma clustering, with most related forms concentrated at very low distances (peak around 0.4-0.5). The morpheme model shows intermediate clustering behavior with a broader intra-lemma distribution, while the word model exhibits the most dispersed intra-lemma distances. All models maintain clear separation between intra-lemma and inter-lemma distances, with inter-lemma distributions consistently shifted toward higher distances (0.7-1.0).

\begin{table}[h]
\centering
\small
\begin{tabular}{lccc}
\toprule
\textbf{Metric} & \textbf{Word} & \textbf{Morph} & \textbf{BPE} \\
\midrule
Separation Ratio & 1.44 & 1.31 & 1.45 \\
Cohesion & Moderate & Moderate & Highest \\
Pattern & Dispersed & Inter. & Concentrated \\
\bottomrule
\end{tabular}
\caption{Allomorph clustering characteristics}
\label{tab:allomorph_stats}
\end{table}

The allomorph clustering analysis provides crucial insight into BPE's apparent morphological similarity advantage. While BPE achieves the highest separation ratio and tightest intra-lemma clustering, this pattern may indicate overly uniform embeddings rather than superior morphological understanding. BPE's concentrated intra-lemma distribution suggests that morphologically related forms are clustered so tightly that fine-grained morphological distinctions may be lost.

This finding aligns with the neighbor rank analysis showing BPE's flatter similarity dropoff patterns. Together, these results suggest that BPE creates embedding spaces where morphologically related words are highly similar to each other, but this comes at the cost of reduced discriminative power and less structured semantic organization. The morpheme model's intermediate clustering behavior may represent a better balance between morphological coherence and semantic discrimination.

\subsubsection{Similarity Distribution Analysis}

\begin{figure}[h]
\centering
\includegraphics[width=\columnwidth]{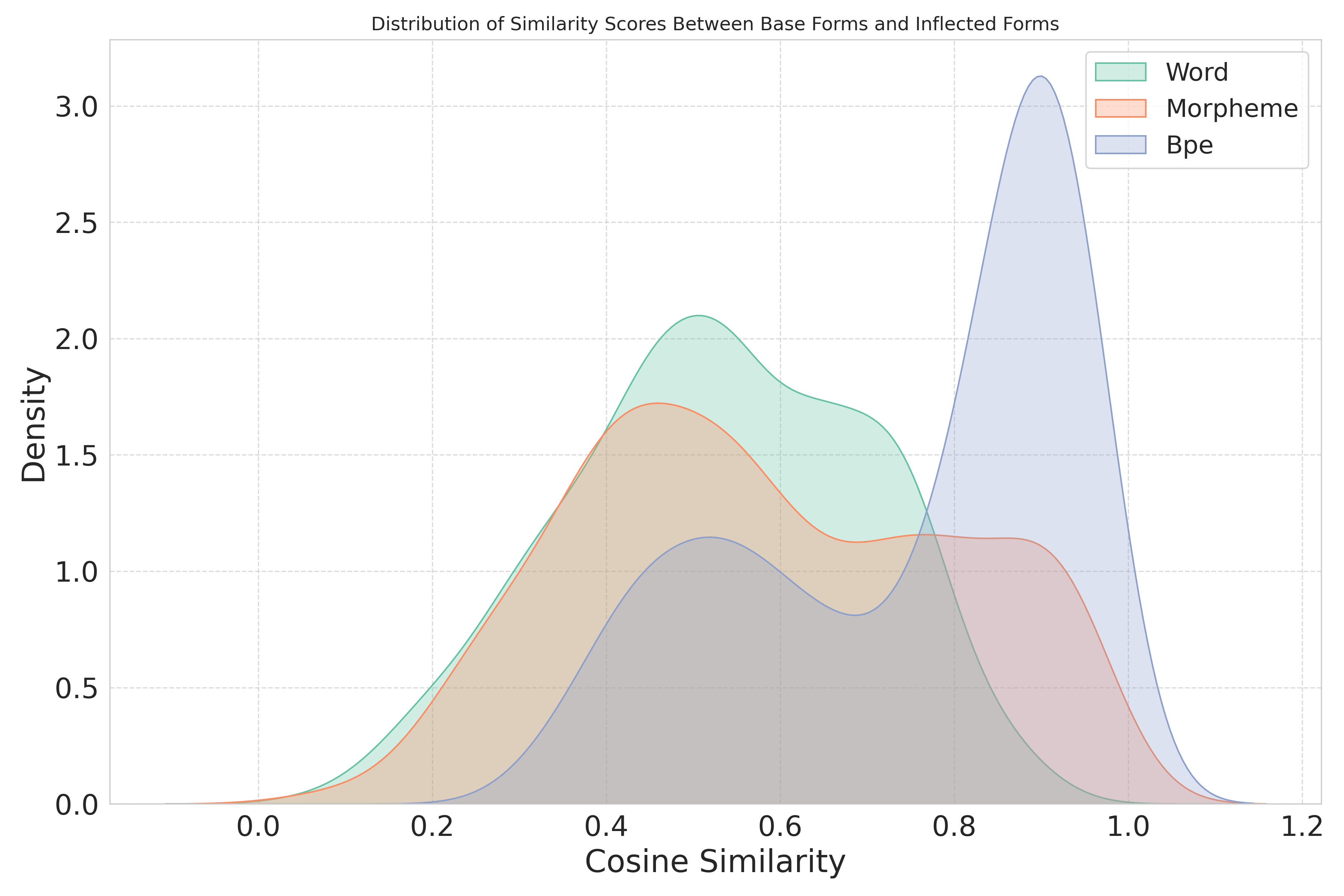}
\caption{Similarity score distributions showing BPE's concentration in high-similarity ranges versus word and morpheme models' broader coverage.}
\label{fig:sim_dist}
\end{figure}

We analyze the distribution of morphological similarity scores to understand each approach's evaluation coverage and bias patterns. Rather than examining only average performance, we investigate how frequently different similarity values occur across all lemma-wordform pairs to identify potential evaluation biases (Figure \ref{fig:sim_dist}). BPE shows extreme concentration in high similarity ranges (55.4\% of pairs achieve 0.8-1.0 similarity) with a sharp peak around 0.9. Word and morpheme models exhibit broader, more balanced distributions centered around 0.4-0.5.

\begin{table}[h]
\centering
\footnotesize
\begin{tabular}{@{}lrrr@{}}
\toprule
\textbf{Range} & \textbf{Word} & \textbf{Morph} & \textbf{BPE} \\
\midrule
0.0-0.4 & 23.4\% & 21.9\% & 3.9\% \\
0.4-0.8 & 72.9\% & 57.0\% & 40.7\% \\
0.8-1.0 & 3.7\% & 21.1\% & 55.4\% \\
\bottomrule
\end{tabular}
\caption{Similarity score distribution by similarity range and tokenization strategy.}
\end{table}

BPE's skewed distribution provides strong evidence for coverage bias. The sharp concentration at high similarities (55.4\% vs 3.7\% for word models) indicates BPE primarily evaluates morphological pairs with straightforward concatenative patterns where compositional vectors work well, such as regular stem-affix combinations. BPE fails to evaluate more challenging morphological relationships involving stem changes, irregular forms, or complex phonological processes that word and morpheme models can handle across the full complexity spectrum. This distribution pattern explains BPE's artificially inflated average similarity scores, as it essentially cherry-picks the morphological cases it can process successfully while being unable to evaluate the challenging cases that would reveal its limitations. The apparent superiority reflects selective evaluation of favorable cases rather than genuine morphological understanding.

\subsection{Vocabulary Characteristics and Overlap Analysis}

We analyze vocabulary sizes and overlap patterns across tokenization approaches to understand their fundamental differences in linguistic unit identification.

The three approaches produce dramatically different vocabulary characteristics. Word-level tokenization generates the largest vocabulary (260,922 tokens), morpheme-based segmentation produces an intermediate vocabulary (162,504 tokens), while BPE creates the smallest vocabulary (2,273 tokens), representing a 115-fold size difference.

Overlap analysis reveals striking patterns: morpheme and word models share substantial convergence (114,639 tokens, 70.5\% of morpheme vocabulary), indicating many morphemes correspond to complete words, particularly for morphologically simple forms. In contrast, morpheme-BPE overlap is minimal (2,034 tokens, 1.3\% of morpheme vocabulary), while BPE-word overlap appears high relative to BPE's small size (2,031 tokens, 89.4\% of BPE vocabulary). This demonstrates that statistical and linguistic approaches identify almost entirely different sets of meaningful units.

\begin{table}[h]
\centering
\tiny
\begin{tabular}{lrcc}
\toprule
\textbf{Approach} & \textbf{Vocab Size} & \textbf{Coverage} & \textbf{Key Overlaps} \\
\midrule
Word-level & 260,922 & 94.3\% & -- \\
Morpheme & 162,504 & 68.7\% & 70.5\% with Word model \\
BPE & 2,273 & 28.6\% & 89.4\% w/ Word model, 1.3\% w/ Morph model \\
\bottomrule
\end{tabular}
\caption{Vocabulary and overlap patterns}
\end{table}

\subsection{Coverage Disparities and Evaluation Bias}

\begin{figure}[h]
\centering
\includegraphics[width=0.45\textwidth]{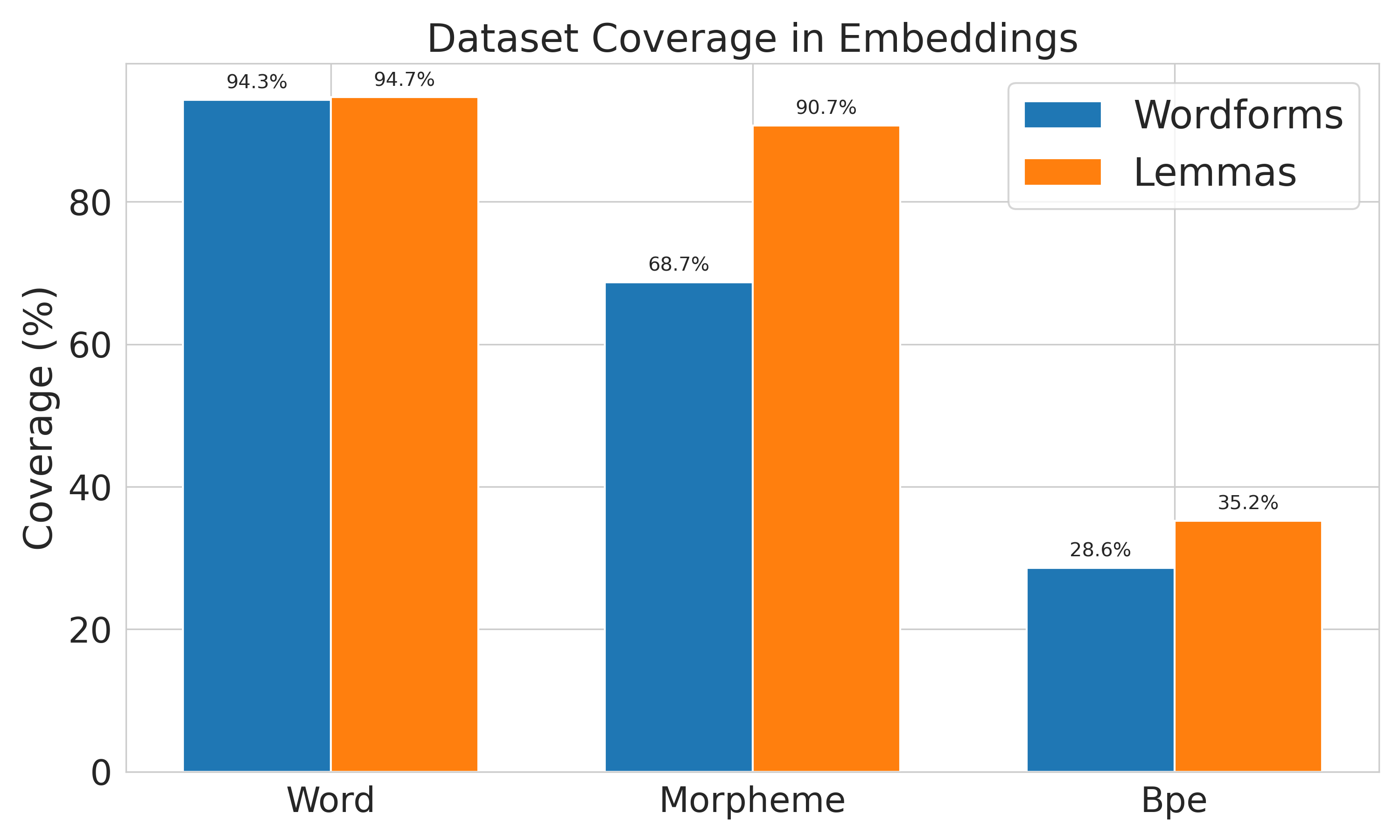}
\caption{Dataset coverage in embeddings showing dramatic differences in evaluation coverage across tokenization approaches}
\label{fig:vocab_coverage}
\end{figure}

The vocabulary differences documented above have profound implications for fair evaluation. Data coverage refers to the percentage of UniMorph lemma-wordform pairs each model can actually evaluate during testing (Figure \ref{fig:vocab_coverage}). For a morphological similarity assessment to be computed, both the lemma and its inflected form must have available embeddings, either through direct vocabulary lookup or through compositional vector construction from subword components.

This coverage analysis reveals severe disparities: BPE can evaluate only 28.6\% of UniMorph test cases, while morpheme models handle 68.7\% and word-level models achieve 94.3\% coverage. These differences introduce systematic evaluation bias, as models are essentially being tested on different subsets of morphological complexity.

BPE's apparent superior performance (0.752 average similarity) may reflect evaluation on a highly selective subset where compositional vector construction succeeds, primarily cases with straightforward concatenative morphology. Meanwhile, word and morpheme models are evaluated across substantially larger and more representative portions of the morphological complexity spectrum. This coverage bias fundamentally compromises the validity of direct performance comparisons and suggests that BPE's morphological similarity advantage may be artifactual rather than substantive.

\section{Conclusion and Future Work}

\subsection{Summary of Key Findings}

Our evaluation of Kurdish tokenization strategies reveals nuanced patterns that challenge simple assumptions about morphological segmentation effectiveness. While BPE initially appears to outperform morpheme-based approaches (0.752 vs 0.583 average morphological similarity), this advantage is fundamentally compromised by severe evaluation coverage limitations. BPE's evaluation coverage is severely limited (28.6\% of test cases) compared to morpheme models (68.7\%) and word models (94.3\%), suggesting its superior performance reflects selection bias toward favorable cases rather than genuine morphological understanding.
The segmentation agreement analysis confirms fundamental divergence between approaches, with only 14.4\% average boundary agreement and 63.6\% zero-agreement cases. Despite processing identical text, morpheme and BPE methods identify almost entirely different meaningful units (1.3\% vocabulary overlap), with BPE producing nearly twice the tokenization density (3.75 vs 1.99 tokens per word).
Critically, multiple embedding quality measures beyond morphological similarity reveal BPE's limitations. The neighbor rank analysis shows BPE creates less structured embedding spaces with flatter similarity dropoff patterns (39.1\% vs 44.2\% for morphemes), while clustering quality metrics favor word and morpheme models. These findings suggest BPE's apparent morphological advantage comes at the cost of overall semantic organization and discriminative power.

\subsection{Broader Implications}

Our results highlight the complexity of evaluating tokenization approaches for morphologically rich languages. The coverage bias phenomenon demonstrates that high similarity scores can be misleading when based on selective evaluation subsets. This finding has broader implications for low-resource language processing, where compositional vector approaches may systematically exclude challenging morphological relationships.

The minimal agreement between statistical and linguistic segmentation approaches (14.4\%) suggests these methods capture complementary rather than competing aspects of linguistic structure. Frequency-based patterns and morphological boundaries may represent different but equally valid perspectives on meaningful unit identification, necessitating hybrid approaches for comprehensive coverage.

The BiLSTM-CRF segmentation results reveal substantial variation across part-of-speech categories (90.2\% for nouns vs 41.7\% for verbs), creating natural experiments for understanding how segmentation quality propagates to embedding performance. This POS-specific variation provides insights into the linguistic complexity that tokenization systems must address.

\subsection{Future Directions}

Future work should focus on developing fair comparison methodologies across tokenization approaches with different coverage characteristics, including restricted evaluation frameworks and coverage-robust similarity metrics. Given the complementary nature of statistical and linguistic approaches, exploring morphologically-constrained BPE algorithms and multi-level frameworks that combine both insights represents a promising direction.
The poor verb segmentation performance (41.7\%) indicates need for specialized Kurdish morphological processing through verb-specific models, phonological change rules, and complex affixation pattern handling. Practical validation through downstream Kurdish NLP tasks would complement intrinsic measures, while investigating tokenization impacts on transformer-based models offers opportunities for Kurdish-specific language model development.
Future advances will emerge from hybrid methods combining different tokenization paradigms' strengths while addressing individual limitations.

\section{Limitations}

The BiLSTM-CRF morphological segmentation shows poor performance on Kurdish verbs (41.7\% accuracy), potentially undermining morpheme-based approaches' effectiveness. However, this may reflect our training methodology rather than inherent constraints. The  BiLSTM-CRF model was trained using random sampling with bootstrap annotation, without explicit control over part-of-speech distribution. Given the model's strong performance on nouns (90.2\%) and adjectives (90.1\%), the poor verb segmentation may reflect insufficient exposure to verbal morphological patterns during training rather than fundamental model limitations. A more balanced training approach with deliberate inclusion of diverse verb forms might significantly improve segmentation performance and, consequently, morpheme-based embedding quality.
The evaluation employs intrinsic similarity measures, which provide controlled assessment of morphological relationship preservation but would benefit from complementary downstream task validation to demonstrate practical applicability. However, comprehensive downstream evaluation is constrained by the limited availability of Kurdish NLP datasets and task-specific resources, a common challenge in low-resource language research.

\bibliographystyle{icml2025}
\bibliography{references}

@article{rudersurevey,
author = {Ruder, Sebastian and Vuli\'{c}, Ivan and S\o{}gaard, Anders},
title = {A survey of cross-lingual word embedding models},
year = {2019},
issue_date = {May 2019},
publisher = {AI Access Foundation},
address = {El Segundo, CA, USA},
volume = {65},
number = {1},
issn = {1076-9757},
url = {https://doi.org/10.1613/jair.1.11640},
doi = {10.1613/jair.1.11640},
journal = {J. Artif. Int. Res.},
month = may,
pages = {569–630},
numpages = {62}
}

@inproceedings{Erdmann2018ComplementarySFA,
    title = "Complementary Strategies for Low Resourced Morphological Modeling",
    author = "Erdmann, Alexander  and
      Habash, Nizar",
    editor = "Kuebler, Sandra  and
      Nicolai, Garrett",
    booktitle = "Proceedings of the Fifteenth Workshop on Computational Research in Phonetics, Phonology, and Morphology",
    month = oct,
    year = "2018",
    address = "Brussels, Belgium",
    publisher = "Association for Computational Linguistics",
    url = "https://aclanthology.org/W18-5806/",
    doi = "10.18653/v1/W18-5806",
    pages = "54--65"
}

@article{bostrom2020byte,
   title={Byte Pair Encoding is Suboptimal for Language Model Pretraining},
   author={Bostrom, Kaj and Durrett, Greg},
   journal={Findings of the Association for Computational Linguistics: EMNLP 2020},
   pages={4617--4624},
   year={2020}
 }

@article{Park2020MorphologyMAA,
  title={Morphology Matters: A Multilingual Language Modeling Analysis},
  author={Hyunji Hayley Park and Katherine J. Zhang and Coleman Haley and K. Steimel and Han Liu and Lane Schwartz},
  journal={Transactions of the Association for Computational Linguistics},
  year={2020},
  volume={9},
  pages={261-276},
  url={https://doi.org/10.1162/tacl\_a\_00365}
}

@inproceedings{cotterell2016morphological,
   title={Morphological Smoothing and Extrapolation of Word Embeddings},
   author={Cotterell, Ryan and Sch{\"u}tze, Hinrich},
   booktitle={Proceedings of the 54th Annual Meeting of the Association for Computational Linguistics},
   pages={1651--1660},
   year={2016}
 }

@inproceedings{sheykhsorani,
    title = "{S}orani {K}urdish versus {K}urmanji {K}urdish: An Empirical Comparison",
    author = "Sheykh Esmaili, Kyumars  and
      Salavati, Shahin",
    editor = "Schuetze, Hinrich  and
      Fung, Pascale  and
      Poesio, Massimo",
    booktitle = "Proceedings of the 51st Annual Meeting of the Association for Computational Linguistics (Volume 2: Short Papers)",
    month = aug,
    year = "2013",
    address = "Sofia, Bulgaria",
    publisher = "Association for Computational Linguistics",
    url = "https://aclanthology.org/P13-2054/",
    pages = "300--305"
}

@inproceedings{luong2013better,
   title={Better Word Representations with Recursive Neural Networks for Morphology},
   author={Luong, Thang and Socher, Richard and Manning, Christopher D},
   booktitle={Proceedings of the Seventeenth Conference on Computational Natural Language Learning},
   pages={104--113},
   year={2013}
 }

@inproceedings{sennrich2016neural,
   title={Neural Machine Translation of Rare Words with Subword Units},
   author={Sennrich, Rico and Haddow, Barry and Birch, Alexandra},
   booktitle={Proceedings of the 54th Annual Meeting of the Association for Computational Linguistics},
   pages={1715--1725},
   year={2016}
 }

@inproceedings{mielke2019kind,
   title={What Kind of Language Is Hard to Language-Model?},
   author={Mielke, Sabrina J and Cotterell, Ryan and Gorman, Kyle and Roark, Brian and Eisner, Jason},
   booktitle={Proceedings of the 57th Annual Meeting of the Association for Computational Linguistics},
   pages={4975--4989},
   year={2019}
 }

@article{ahmadi2019tokenization,
   title={Tokenization for Low-Resource Languages: A Case Study on {K}urdish},
   author={Ahmadi, Sina and Wurm, Mark Lee},
   journal={arXiv preprint arXiv:1909.07739},
   year={2019}
 }

@inproceedings{gerz2018relation,
   title={On the Relation between Linguistic Typology and (Limitations of) Multilingual Language Modeling},
   author={Gerz, Daniela and Vulic, Ivan and Ponti, Edoardo Maria and Naradowsky, Jason and Reichart, Roi and Korhonen, Anna},
   booktitle={Proceedings of the 2018 Conference on Empirical Methods in Natural Language Processing},
   pages={316--327},
   year={2018}
 }

@inproceedings{kudo2018subword,
 title={Subword regularization: Improving neural network translation models with multiple subword candidates},
 author={Kudo, Taku},
 booktitle={Proceedings of the 56th Annual Meeting of the Association for Computational Linguistics (Volume 1: Long Papers)},
 pages={66--75},
 year={2018}
}

@inproceedings{smit-etal-2014-morfessor,
    title = "{M}orfessor 2.0: Toolkit for statistical morphological segmentation",
    author = {Smit, Peter  and
      Virpioja, Sami  and
      Gr{\"o}nroos, Stig-Arne  and
      Kurimo, Mikko},
    editor = "Wintner, Shuly  and
      Tadi{\'c}, Marko  and
      Babych, Bogdan",
    booktitle = "Proceedings of the Demonstrations at the 14th Conference of the {E}uropean Chapter of the Association for Computational Linguistics",
    month = apr,
    year = "2014",
    address = "Gothenburg, Sweden",
    publisher = "Association for Computational Linguistics",
    url = "https://aclanthology.org/E14-2006/",
    doi = "10.3115/v1/E14-2006",
    pages = "21--24"
}

@inproceedings{schuster2012japanese,
 title={Japanese and korean voice search},
 author={Schuster, Mike and Nakajima, Kaisuke},
 booktitle={2012 IEEE International Conference on Acoustics, Speech and Signal Processing (ICASSP)},
 pages={5149--5152},
 year={2012},
 organization={IEEE}
}

@inproceedings{rehurek_lrec,
      title = {{Software Framework for Topic Modelling with Large Corpora}},
      author = {Radim {\v R}eh{\r u}{\v r}ek and Petr Sojka},
      booktitle = {{Proceedings of the LREC 2010 Workshop on New
           Challenges for NLP Frameworks}},
      pages = {45--50},
      year = 2010,
      month = May,
      day = 22,
      publisher = {ELRA},
      address = {Valletta, Malta},
      language={English}
}

@inproceedings{mikolov2013distributed,
  title={Distributed representations of words and phrases and their compositionality},
  author={Mikolov, Tomas and Sutskever, Ilya and Chen, Kai and Corrado, Greg S and Dean, Jeff},
  booktitle={Advances in neural information processing systems},
  pages={3111--3119},
  year={2013}
}

@inproceedings{kudo2018sentencepiece,
 title={SentencePiece: A simple and language independent subword tokenizer and detokenizer for neural text processing},
 author={Kudo, Taku and Richardson, John},
 booktitle={Proceedings of the 2018 Conference on Empirical Methods in Natural Language Processing: System Demonstrations},
 pages={66--71},
 year={2018}
}

@article{esmaili2013sorani,
   title={{S}orani {K}urdish versus {K}urmanji {K}urdish: An Empirical Comparison},
   author={Esmaili, Kyumars Sheykh and Salavati, Shahin},
   journal={Proceedings of the 51st Annual Meeting of the Association for Computational Linguistics},
   pages={300--305},
   year={2013}
 }

@article{hassani2018blark,
   title={{BLARK} for multi-dialect languages: towards the {K}urdish {BLARK}},
   author={Hassani, Hossein},
   journal={Language Resources and Evaluation},
   volume={52},
   number={2},
   pages={625--644},
   year={2018}
 }

@article{ahmadi2020towards,
   title={Towards electronic lexicography for the {K}urdish language},
   author={Ahmadi, Sina and Hassani, Hossein and McCrae, John P},
   journal={Proceedings of the 7th Workshop on NLP for Similar Languages, Varieties and Dialects},
   pages={9--18},
   year={2020}
 }

@article{Ahmadi2020TowardsFM,
  title={Towards Finite-State Morphology of Kurdish},
  author={Sina Ahmadi and Hossein Hassani},
  journal={ArXiv},
  year={2020},
  volume={abs/2005.10652},
  url={https://api.semanticscholar.org/CorpusID:218763544}
}

@inproceedings{ahmadi-2020-tokenization,
    title = "A Tokenization System for the {K}urdish Language",
    author = "Ahmadi, Sina",
    editor = {Zampieri, Marcos  and
      Nakov, Preslav  and
      Ljube{\v{s}}i{\'c}, Nikola  and
      Tiedemann, J{\"o}rg  and
      Scherrer, Yves},
    booktitle = "Proceedings of the 7th Workshop on NLP for Similar Languages, Varieties and Dialects",
    month = dec,
    year = "2020",
    address = "Barcelona, Spain (Online)",
    publisher = "International Committee on Computational Linguistics (ICCL)",
    url = "https://aclanthology.org/2020.vardial-1.11/",
    pages = "114--127"
}

@article{creutz2005unsupervised,
  title={Unsupervised morpheme segmentation and morphology induction from text corpora using {M}orfessor 1.0},
  author={Creutz, Mathias and Lagus, Krista},
  journal={Helsinki University of Technology},
  year={2005}
}

@article{MAHMUDI2021101222,
title = {Automated grapheme-to-phoneme conversion for Central Kurdish based on optimality theory},
journal = {Computer Speech \& Language},
volume = {70},
pages = {101222},
year = {2021},
issn = {0885-2308},
doi = {https://doi.org/10.1016/j.csl.2021.101222},
url = {https://www.sciencedirect.com/science/article/pii/S0885230821000292},
author = {Aso Mahmudi and Hadi Veisi}
}

@inproceedings{lafferty2001conditional,
  title={Conditional random fields: Probabilistic models for segmenting and labeling sequence data},
  author={Lafferty, John and McCallum, Andrew and Pereira, Fernando CN},
  booktitle={Proceedings of the Eighteenth International Conference on Machine Learning},
  pages={282--289},
  year={2001}
}

@book{thackston2006sorani,
  title={Sorani Kurdish: A Reference Grammar with Selected Readings},
  author={Thackston, Wheeler M.},
  publisher={Harvard University Press},
  year={2006},
  address={Cambridge, MA}
}

@book{manning1999foundations,
  title={Foundations of Statistical Natural Language Processing},
  author={Manning, Christopher D and Schütze, Hinrich},
  year={1999},
  publisher={MIT Press}
}

@inproceedings{pimentel-ryskina-etal-2021-sigmorphon,
    title = "SIGMORPHON 2021 Shared Task on Morphological Reinflection: Generalization Across Languages",
    author = "Pimentel, Tiago  and
      Ryskina, Maria  and
      Mielke, Sabrina J.  and
      Wu, Shijie  and
      Chodroff, Eleanor  and
      Leonard, Brian  and
      Nicolai, Garrett  and
      Ghanggo Ate, Yustinus  and
      Khalifa, Salam  and
      Habash, Nizar  and
      El-Khaissi, Charbel  and
      Goldman, Omer  and
      Gasser, Michael  and
      Lane, William  and
      Coler, Matt  and
      Oncevay, Arturo  and
      Montoya Samame, Jaime Rafael  and
      Silva Villegas, Gema Celeste  and
      Ek, Adam  and
      Bernardy, Jean-Philippe  and
      Shcherbakov, Andrey  and
      Bayyr-ool, Aziyana  and
      Sheifer, Karina  and
      Ganieva, Sofya  and
      Plugaryov, Matvey  and
      Klyachko, Elena  and
      Salehi, Ali  and
      Krizhanovsky, Andrew  and
      Krizhanovsky, Natalia  and
      Vania, Clara  and
      Ivanova, Sardana  and
      Salchak, Aelita  and
      Straughn, Christopher  and
      Liu, Zoey  and
      Washington, Jonathan North  and
      Ataman, Duygu  and
      Kiera{\'s}, Witold  and
      Woli{\'n}ski, Marcin  and
      Suhardijanto, Totok  and
      Stoehr, Niklas  and
      Nuriah, Zahroh  and
      Ratan, Shyam  and
      Tyers, Francis M.  and
      Ponti, Edoardo M.  and
      Aiton, Grant  and
      Hatcher, Richard J.  and
      Prud'hommeaux, Emily  and
      Kumar, Ritesh  and
      Hulden, Mans  and
      Barta, Botond  and
      Lakatos, Dorina  and
      Szolnok, G{\'a}bor  and
      {\'A}cs, Judit  and
      Raj, Mohit  and
      Yarowsky, David  and
      Cotterell, Ryan  and
      Ambridge, Ben  and
      Vylomova, Ekaterina",
    booktitle = "Proceedings of the 18th SIGMORPHON Workshop on Computational Research in Phonetics, Phonology, and Morphology",
    month = aug,
    year = "2021",
    address = "Online",
    publisher = "Association for Computational Linguistics",
    url = "https://aclanthology.org/2021.sigmorphon-1.25",
    doi = "10.18653/v1/2021.sigmorphon-1.25",
    pages = "229--259"
}

@article{Huang2015BidirectionalLM,
  title={Bidirectional LSTM-CRF Models for Sequence Tagging},
  author={Zhiheng Huang and Wei Xu and Kai Yu},
  journal={ArXiv},
  year={2015},
  volume={abs/1508.01991},
  url={https://api.semanticscholar.org/CorpusID:12740621}
}

@article{Ahmadi2021HunspellFS,
  title={Hunspell for Sorani Kurdish Spell Checking and Morphological Analysis},
  author={Sina Ahmadi},
  journal={ArXiv},
  year={2021},
  volume={abs/2109.06374},
  url={https://api.semanticscholar.org/CorpusID:237503500}
}

@article{asosoftcorpuspaper,
    author = {Veisi, Hadi and MohammadAmini, Mohammad and Hosseini, Hawre},
    title = "Toward {K}urdish language processing: Experiments in collecting and processing the {AsoSoft} text corpus",
    journal = {Digital Scholarship in the Humanities},
    volume = {35},
    number = {1},
    pages = {176-193},
    year = {2019},
    month = {02},
    issn = {2055-7671},
    doi = {10.1093/llc/fqy074},
    url = {https://doi.org/10.1093/llc/fqy074}
}

\newpage
\section*{Appendix} 
\label{appendix}

\subsection*{Corpus Preprocessing and Normalization}
\label{prep}

The Kurdish corpus required extensive preprocessing to address orthographic inconsistencies and dialectal variations inherent in Kurdish text collections. This section includes more details about the normalization procedures applied to ensure consistent tokenization and analysis.

\textbf{Character Repetition Normalization:} User-generated content frequently contained repeated characters for emphasis (e.g., \ku{سڵاووووو} for emphasis). We implemented a normalization rule that reduced any character sequence repeated more than three times to exactly three repetitions, thus \ku{سڵاووووو} became \ku{سڵاووو}. This approach balanced noise reduction with preservation of linguistic nuance, as triple repetition often carries semantic meaning in Kurdish.

\textbf{Script and Character Standardization:} Sorani Kurdish exhibits pervasive orthographic inconsistencies in some characters, particularly with characters \ku{ە} and \ku{ه}, which represent the same phoneme but are used differently based on context, keyboard layout, and individual preference. These variations often involve different Unicode code points, creating artificial vocabulary inflation. We developed a custom character replacement system that mapped these variants to canonical forms, ensuring consistent representation throughout the corpus.

\textbf{Zero-Width Non-Joiner (ZWNJ) Handling:} Arabic-based script languages like Kurdish frequently contain zero-width non-joiner characters that alter character joining behavior without visible effect, disrupting consistent tokenization. Different keyboard layouts encode ZWNJ differently. Some through dedicated keys, others through key combinations—leading to inconsistent usage patterns. Users sometimes substitute full spaces or omit the character entirely, creating tokenization ambiguity. We implemented corpus-wide ZWNJ regularization through text-level normalization routines that either removed or standardized these characters based on contextual appropriateness.

\textbf{Quality Filtering:} We applied AsoSoft's text normalization method from their Python library in initial preprocessing stages. Subsequently, sentences shorter than 5 characters were removed to eliminate fragments and malformed entries. Additionally, lines lacking valid Kurdish characters (identified using Unicode ranges for Kurdish script) were filtered out to ensure corpus linguistic consistency.

\textbf{Dialectal Filtering:} The corpus contained multi-dialectal interference from Kurmanji Kurdish and Persian sources. We developed character profile analysis to identify sentences deviating from Sorani Kurdish orthographic norms, filtering out content that exhibited non-Sorani characteristics. This process helped maintain dialectal consistency while preserving corpus size.

\textbf{Format Standardization:} Final preprocessing involved segmenting documents into sentence-per-line and word-per-line formats to support various downstream tasks including morphological segmentation training and embedding model development. This dual-format approach enabled flexible corpus utilization across different experimental requirements.

\subsection*{Implementation Details}

All experiments were conducted using Python 3.9.6 with PyTorch 1.12.0 for neural model implementation. The BiLSTM-CRF morphological segmenter was trained using the torchcrf library (1.1.0) with Adam optimizer and learning rate scheduling. word2vec models were trained using Gensim 4.2.0 with identical hyperparameters across tokenization approaches to ensure fair comparison.

BPE tokenization employed the Hugging Face tokenizers library (0.15.2) with vocabulary size set to 2,280 tokens based on empirical morpheme count analysis. The tokenizer was trained on the full corpus before applying to sample data for embedding training. Morphological segmentation used our BiLSTM-CRF model with 3-layer bidirectional LSTM (hidden size 128) and CRF output layer.

\subsection*{Experimental Configuration}

word2vec training employed skip-gram architecture with negative sampling (5 negative samples), vector dimensions of 150, minimum count threshold of 5, and 10 training epochs. Window sizes were adjusted proportionally based on tokenization density: word-level (5), morpheme-based (10), and BPE (19). All models used identical random seeds (42) for reproducibility.

The evaluation corpus comprised 1.5 million sentences sampled from a larger Kurdish text collection \citep{asosoftcorpuspaper}. UniMorph evaluation used 996 Kurdish lemma-wordform pairs with morphological feature annotations. Coverage analysis and similarity computations employed cosine similarity with compositional vector generation for out-of-vocabulary terms through subword averaging.

\subsection*{Computational Resources}

Experiments were conducted on systems with NVIDIA RTX 3080 GPUs and 32GB RAM. BiLSTM-CRF training required less than an hour, while Word2Vec training ranged from 4 minutes (word-level) to 25 minutes (BPE) depending on vocabulary size and tokenization complexity.

All code and trained models will be available to facilitate reproduction and extension of these results.

\end{document}